\documentclass[letter, 10pt]{IEEEtran}

\usepackage{amsmath}
\usepackage{amssymb}
\usepackage{graphicx}
\usepackage{booktabs}
\usepackage{multirow}
\usepackage{xcolor}
\usepackage{color}
\usepackage[]{algorithm2e}
\usepackage{algpseudocode}
\usepackage{pifont}
\usepackage{siunitx}
\usepackage{makecell}
\usepackage{colortbl}
\usepackage{pgfplots}
\usepackage{balance}
\usepackage{flushend}
\usepackage{pbalance}
\usepackage{hyperref}  % must be loaded last
\usepackage[normalem]{ulem}
\usepackage{soul}
\usepackage{verbatim}

\usepgfplotslibrary{groupplots}
\newcommand{\degree}{^\circ}

\newcommand{\smarttd}{SMART\nobreakdash-3D}
\newcommand{\ignore}[1]{}
\definecolor{bestgreen}{RGB}{220,245,220}
\definecolor{benchblue}{RGB}{59,130,246}
\definecolor{benchorange}{RGB}{245,158,11}
\definecolor{benchgreen}{RGB}{16,185,129}
\definecolor{benchgray}{RGB}{107,114,128}

 \definecolor{pink}{rgb}{1, 0, 1}
 \definecolor{orange}{rgb}{1, 0.7529, 0}
 \definecolor{darkgreen}{rgb}{0, 0.8, 0}

\pgfplotsset{compat=1.18}
\pgfplotsset{
benchmark/.style={
    ybar,
    bar width=6pt,
    width=\linewidth,
    height=5.1cm,
    ymin=0,
    xmin=-0.5,
    xmax=6.5,
    xtick={0,1,2,3,4,5,6},
    xticklabels={building,office,warehouse,hospital,airport,campus,university},
    axis x line*=bottom,
    axis y line*=left,
    ymajorgrids=true,
    grid style={dashed,gray!30},
    tick label style={font=\scriptsize},
    label style={font=\small},
    legend style={font=\scriptsize, at={(0.5,0.97)}, anchor=north, legend columns=2, draw=none, fill=white, fill opacity=0.85, text opacity=1},
    x tick label style={rotate=28, anchor=east, font=\scriptsize},
}
}

\title{\Huge 
PECMAN: Perception-enabled Collaborative Multi-Agent Navigation in Unknown Environments 
}

\author{Tianchonghui Fang$^{1}$, Shaunak Roy$^{1}$ and Shalabh Gupta$^{1}$%
\thanks{$^{1}$ T. Fang (email: tianchonghui.fang@uconn.edu), S. Roy (email: shaunak.roy@uconn.edu) and S. Gupta (email: shalabh.gupta@uconn.edu) are with the Department of Electrical and Computer Engineering, University of Connecticut, Storrs, CT 06269, USA.
Corresponding author: S. Gupta.
}}

\begin{document}

\maketitle
\thispagestyle{empty}

%%%%%%%%%%%%%%%%%%%%%%%%%%%%%%%%%%%%%%%%%%%%%%%%%%%%%%%%%%%%%%%%%%%%%%%%%%%%%%%%
\begin{abstract}

Most path planners assume fully known, static environments, assumptions that fail when robots navigate in dynamic and partially observable environments. \smarttd{} addresses these issues by real-time replanning, where it morphs the underlying RRT* tree whenever new obstacles or structures are discovered in the environment. Instead of rebuilding the tree entirely from scratch, SMART-3D prunes invalid nodes and edges and subsequently repairs the disjoint subtrees at hot-nodes to find a new path, thus providing high computational efficiency for real-time adaptability. We extend \smarttd{} to perception-enabled collaborative multi-agent navigation (PECMAN) in unknown environments. PECMAN is built upon distributed tree morphing and shared perception strategies, where each agent reacts to environmental changes and morphs its respective tree to replan its path, while simultaneously broadcasting newly discovered structures to other agents, thus enabling them to proactively replan even in areas that have not yet been explored by them. This approach reduces redundant reactions and unnecessary replannings of the agents due to improved situational awareness. The performance of PECMAN was evaluated by $28{,}000$ multi-agent simulations on seven $2$D scenarios with different case studies. The results show that PECMAN achieves up to $52\%$ reduction in the team-completion time, while maintaining $\sim100\%$ success rates. Finally, PECMAN was tested by real experiments on two autonomous robots in a building environment. 
\end{abstract}

%%%%%%%%%%%%%%%%%%%%%%%%%%%%%%%%%%%%%%%%%%%%%%%%%%%%%%%%%%%%%%%%%%%%%%%%%%%%%%%%
\section{INTRODUCTION}

Recent years have witnessed an unprecedented increase in the applications of unmanned vehicles (e.g., unmanned air vehicles (UAVs) and unmanned underwater vehicles (UUVs)) including marine life exploration~\cite{Katzschmann2018}\cite{Yuh2011_Applications_marinerobot}, seafloor mapping~\cite{shen2022ct,palomeras2018}, coverage planning~\cite{Cstar2026,song2018}, ocean sensing~\cite{Somers2016}, agriculture~\cite{Moradi22, agriculture13020354}, bridge monitoring~\cite{PANIGATI2025106101}, oil spill cleaning~\cite{SongQiu2014,Kumar2020}, human-robot interaction~\cite{YangWilson2023}, target tracking~\cite{Shojaei2017, Hare2018, Hare2020},  mine countermeasures~\cite{Acar2003, mukherjee2011symbolic}, energy-constrained exploration~\cite{shen2020}, terrain mapping~\cite{Fabricio24}, and surveillance~\cite{MGRW11}.

\subsection{Motivation}
In most complex environments, safe and reliable navigation~\cite{WG2025} becomes challenging because robots have to navigate around unknown static (e.g., walls and rocks) and dynamic (e.g., pedestrians and other robots) obstacles. In this regard, a recent algorithm, called Self Morphing Adaptive Replanning Tree (SMART)~\cite{Shen_SMART2023}, introduced an efficient mechanism for real-time replanning when dynamic obstacles enter the robot's Local Reaction Zone (LRZ). \smarttd{}~\cite{agrawal2025smart3d} extended SMART to $3$D workspaces. Instead of rebuilding the tree entirely from scratch, it morphs the current tree by first pruning the affected nodes and edges, and then repairing the resulting disjoint subtrees by making local reconnections at hot-nodes (nodes adjacent to different subtrees), to find a new path. This local morphing provides high computational efficiency for real-time adaptability. \smarttd{} is validated on a single robot in dynamic, partially observable environments, where obstacles are discovered online through onboard sensing. While \smarttd{} handles unknown and dynamic environments for a single robot, its multi-robot deployment introduces several inter-agent coordination and information-sharing challenges~\cite{song2020care}. For example, due to a lack of coordination and information sharing strategy, it is possible that i) multiple agents decide to go towards the same narrow corridor and face a deadlock, or ii) each agent independently re-discovers the same structures that have already been observed by neighboring  agents, thus wasting resources and delaying planning. To address the above problems, this paper extends the single-agent adaptive navigation strategy of \smarttd{} to develop a novel algorithm called perception-enabled collaborative multi-agent navigation (PECMAN).

\subsection{Related Work}
Three lines of multi-agent research are relevant.
\textbf{Multi-Agent Path Finding (MAPF).}
MAPF~\cite{sharon2015cbs,okumura2022pibt,okumura2023lacam} methods are scalable for coordinating hundreds of agents on known graphs with synchronized discrete timesteps. H\"onig et al.~\cite{honig2019persistent} note that both these assumptions break down on physical robots in real environments. Conflict-based Search (CBS)~\cite{sharon2015cbs} finds optimal solutions, but it suffers from computational explosion in narrow corridors~\cite{li2019symmetry}. Priority Inheritance with Backtracking (PIBT)~\cite{okumura2022pibt} scales to thousands of agents via priority inheritance. Lazy Constraints Addition search for MAPF  (LaCAM)~\cite{okumura2023lacam} adds lazy constraints to the search problem. All the above methods require a known grid and discrete timesteps. Lifelong MAPF~\cite{li2021lifelong} handles persistent task streams but still needs a known graph. In contrast, PECMAN handles unknown continuous environments, a regime that the above methods do not address.

\textbf{Sampling-based replanning.}
Sampling-based multi-agent planning~\cite{cap2013marrt,solovey2016drrt} methods handle continuous workspaces using RRT-based exploration, but some of these methods assume that the environment is known. Some methods can also handle dynamic obstacles. Dynamic RRT (DRRT)~\cite{ferguson2006replanning} prunes the invalid tree-structures due to dynamic obstacles and regrows the tree to replan the path. RRTX~\cite{otte2016rrtx} replans by propagating the cost changes globally. Extended RRT (ERRT)~\cite{bruce2002real} grows the tree using a bias toward the old plan. These methods require significant modifications to the tree structure every time a change occurs. On the other hand, \smarttd{}~\cite{agrawal2025smart3d} prunes invalid nodes and edges in a local Critical Pruning Region (CPR) and reconnects disjoint subtrees via hot-nodes, thus preserving most of the tree structure. PECMAN keeps the local prune and repair properties of SMART-3D and builds a multi-agent coordination layer on top.

\textbf{Reactive coordination.}
Optimal Reciprocal Collision Avoidance (ORCA)~\cite{berg2011reciprocal} computes pairwise velocity constraints. It operates in open continuous spaces but can face deadlocks in narrow corridors where no mutually feasible velocity exists. Dynamic Window Approach (DWA)~\cite{fox1997dwa} has the same limitation. In contrast, PECMAN introduces a Global King/flee/push coordination layer to resolve such conflicts by sequencing robots through priority inheritance, where the King agent passes first while the others flee, with the RRT* tree providing fallback paths after displacement.

\subsection{Contributions} 
This paper develops a novel algorithm called PECMAN for collaborative multi-agent navigation in unknown environments. The key contributions of this paper are as follows:

\begin{enumerate}
\item \textbf{Shared perception.} PECMAN enables shared perception in unknown environments. This means that if any agent discovers a static structure (e.g., a wall) in the environment using its LiDAR, it broadcasts this information to all other agents who, upon receiving such early notice, react accordingly and update their plans proactively before detecting this structure themselves. This can significantly reduce the trajectory lengths and times of these agents due to early planning and by not traveling in the wrong direction in the unknown realm. 

\item \textbf{Distributed tree morphing.} PECMAN scales the single agent tree morphing (i.e., prune-and-repair) strategy of SMART-3D to distributed tree morphing for a team of robots. In this distributed strategy, each agent morphs its own RRT* tree in response to a) nearby dynamic obstacles and b) static obstacles discovered either by its own sensor or based on the information received from other agents. In case b), it permanently deletes the tree-portions that are blocked by static obstacles. 
   
\item \textbf{Narrow corridor coordination.} PECMAN introduces a Global King Priority layer to resolve deadlocks at narrow corridors through flee/push behaviors and multi-directional retreat without the need to modify any planning trees. This mechanism is similar to PIBT's priority inheritance~\cite{okumura2022pibt} but in continuous coordinates rather than on a grid.
\end{enumerate}

The performance of PECMAN was validated on seven procedurally generated $2$D scenarios, ranging from $32$\,m$\times$$32$\,m to $192$\,m $\times$ $192$\,m  with up to $4$ agents and $30$ pedestrians, totaling $28{,}000$  multi-agent trials, considering two different repair strategies and two different sensing modes: shared and independent. Finally, PECMAN was tested by real experiments on two autonomous robots in a building environment. 
%%%%%%%%%%%%%%%%%%%%%%%%%%%%%%%%%%%%%%%%%%%%%%%%%%%%%%%%%%%%%%%%%%%%%%%%%%%%%%%%

\section{The PECMAN Algorithm}

\subsection{Problem formulation}

Consider a set $\mathcal{R}=\{\mathcal{R}_1,\mathcal{R}_2,...\mathcal{R}_N\}$ of $N \in \mathbb{Z}_{+}$ agents operating in a workspace $\mathcal{W} \subset \mathbb{R}^2$ with unknown static structures $\mathcal{O}_{\text{st}}\subset \mathcal{W}$ (e.g., walls) and a set $\mathcal{O}_{\text{dy}}=\{\mathcal{O}_1,\mathcal{O}_2,...\mathcal{O}_M\}$  of $M \in \mathbb{Z}_{+}$ dynamic obstacles (e.g., pedestrians). An agent $\mathcal{R}_i \in \mathcal{R}$ has a position $\mathbf{x}_i(t) \in \mathbb{R}^2$, radius $r_i \in \mathbb{R}_+$, speed $v_i \in \mathbb{R}_+$, goal $\mathbf{g}_i \in \mathbb{R}^2$, and LiDAR range $r_s \in \mathbb{R}_+$. It maintains a map $\hat{\mathcal{M}}_i(t) \subseteq \mathcal{O}_{\text{st}}$ and constantly updates it based on newly discovered structures. At $t{=}0$, $\hat{\mathcal{M}}_i(0) = \emptyset$. Furthermore, each agent maintains an RRT* tree $\mathcal{T}_i(t)$, that is constantly morphed based on the latest information about static and dynamic obstacles. At $t{=}0$, $\mathcal{T}_i(0)$ is built using the initial (partial) knowledge about static structures. Finally, each agent also maintains a plan $P_i$, which is the path to $\mathbf{g}_i$. The path is constantly replanned using the updated tree $\mathcal{T}_i(t)$. In addition, each agent $\mathcal{R}_i$ also receives information $\mathcal{I}_{i'}(t)$ from the other agents $\mathcal{R}_{i'}\in \mathcal{R}, i'\neq i$, including their current positions $\mathbf{x}_{i'}(t)$ and the portions of the map $m_{i'}\subseteq \mathcal{O}_{\text{st}}$ that have been newly discovered by them.

\textbf{Objective:} All agents reach their respective goals successfully, avoiding all static and dynamic obstacles, while minimizing $\max_i T_i^{\text{goal}}$, where $T_i^{\text{goal}}$ is the time taken by agent $\mathcal{R}_i$ to reach its goal $\mathbf{g}_i$.

\subsection{Algorithm phases}

The system runs a 4-phase loop every time frame ($\Delta t {=} 0.05$\,s). These phases are described below. 

\vspace{6pt}
\subsubsection{Phase 1: LiDAR Scan}\label{sec:phase1}
In order to map the environment, each agent $\mathcal{R}_i$ performs a $360\degree$ scan of the surroundings using $180$ sensor rays of range $r_s$. This scan detects both static and dynamic obstacles within the sensing range but is unable to detect occluded items behind other obstacles. If it detects any new portions of static structures $m_{i}\subseteq \mathcal{O}_{\text{st}}$, then it updates its own map as $\hat{\mathcal{M}}_{i}(t)\leftarrow\hat{\mathcal{M}}_{i}(t)\cup m_{i}$.

\vspace{6pt}
\subsubsection{Phase 2: Shared Perception}
\label{sec:phase2}
In this phase, each agent broadcasts information about newly discovered structures to a central coordinator, which in turn broadcasts it to all other agents. As such, if agent $\mathcal{R}_i$ receives information from any other agent $\mathcal{R}_{i'}$, then it updates its map as $\hat{\mathcal{M}}_i(t) \leftarrow \hat{\mathcal{M}}_i(t) \cup m_{i'}$, $\forall i'$. This process is repeated for all agents $\mathcal{R}_i\in\mathcal{R}$, until all maps are synchronized. On the other hand, in independent operation mode, each agent relies solely on the map generated by its own sensor for planning and does not receive information from other agents. The advantage of shared discovery is that each agent can make proactive planning decisions, even for regions that it has never scanned using its own sensors, thus utilizing the potential to save significant travel time. For example, if the goal is behind an unknown wall that has a door on its left, then due to the lack of information, the agent would travel in a straight line to the goal and not towards the door until realizing it later upon detecting the wall.  

\vspace{6pt}
\subsubsection{Phase 3: Distributed Tree Repair}
Each agent performs its tree pruning and repair when a) its map is updated to reveal new static structures and/or b) it detects dynamic obstacles. 

$\bullet$ \textbf{Tree morphing for static structures}- To accommodate new static structures, we describe four strategies as follows.
\begin{itemize}
\item [1.]\textit{Full tree rebuild.} In this strategy, a new tree is built from scratch after discarding the previous tree (15{,}000-nodes). This guaranties an optimal RRT$^*$ tree, but costs $\sim40$\,ms per rebuild and expends significant time in replanning. For example, on the campus scenario (${\sim}$300 walls), this means ${\sim}$140 rebuilds per run, totaling ${\sim}$5.6\,s of replanning overhead. Furthermore, this could be orders of magnitude slower on an edge device and does not scale with the scenario size. We use this as a baseline.

\item [2.] \textit{Eager.} In this strategy, all edges $E$ of the tree are scanned, then the edges that intersect with the newly mapped structure are permanently pruned, and finally, local reconnections are made via hot-nodes within the Local Search Radius (LSR) of radius $80$\,m. This preserves most of the tree structure and costs ${\sim}$5\,ms per replanning.  Algorithm~\ref{alg:wall_repair} describes the Eager strategy.

\item [3.] \textit{Lazy Eager (LE).} In this strategy, if the current path is blocked by the new structure, then Eager is triggered; otherwise, replanning is skipped. 
This prevents unnecessary replannings if the new structures dont obstruct the path, thus saving significant replanning time.

\item [4.] \textit{Swift.} This strategy is similar to Lazy Eager except that only the edges near the current path are scanned, and pruned if invalid, followed by tree repair and path search. 

\end{itemize}

\vspace{0pt}
Lazy Eager and Swift strategies are both computationally cheaper; however, they can leave invalid edges elsewhere in the tree. This could cause problems during other replanning incidents or when an agent is pushed towards invalid edges by another robot during King coordination, resulting in a full rebuild. In multi-agent mode with shared perception, such invalid edges can accumulate fast, as new structures are discovered by remote agents.

\RestyleAlgo{ruled}
\LinesNumbered
\begin{algorithm}[t]
%\footnotesize
%\begin{algorithmic}[1]
\KwIn{Tree $\mathcal{T}$, new wall $w$.}
 EdgeValidityCheck($\mathcal{T}$, $w$); \\
 $\mathcal{E}_{\text{bad}} \leftarrow \{e \in \mathcal{T} \mid e \text{ intersects } w\}$ \\
\For{each $e \in \mathcal{E}_{\text{bad}}$}
   {
   Prune $e$; {\ \ \textit{tree fragments into subtrees}} }
$\mathcal{H} \leftarrow$ HotNodeSearch($\mathcal{T}$, $\text{LSR}$)\\
\For{$h \in \mathcal{H}$}
{
    \If{reconnection through $h$ valid} 
        { 
        Reconnect,\\ 
        Rewiring cascade.
        }
}
\textbf{return} BuildPath($\mathcal{T}$). 
%\end{algorithmic}
\caption{Eager}
\label{alg:wall_repair}
\end{algorithm}
\setlength{\textfloatsep}{3pt}

\vspace{3pt}
$\bullet$ \textbf{Tree morphing for dynamic obstacles}-
This is unchanged from the original \smarttd{} algorithm: compute obstacle hazard zones OHZs for each dynamic obstacle, prune invalid nodes and edges inside OHZs of all dynamic obstacles intersecting with the agent's LRZ, reconnect via hot-nodes to morph the tree, and finally find a new path to the goal. 

\vspace{6pt}
\subsubsection{Phase 4: Narrow Corridor Coordination}
\label{sec:king}
This layer resolves deadlocks in narrow corridors without morphing the planning tree of any agent by operating at the control level. 

\textbf{King selection.} The king is the highest-priority active agent. The priorities could be assigned based on different factors, such as task criticality, distance to goal, battery level, and health status. When the king reaches its goal, the next highest priority agent becomes the king. The same priority ties are handled by random assignment. 

\textbf{King behavior.} The king follows its path through the narrow corridor, ignoring all other agents whose paths also pass through the corridor. If a collision situation arises, the king pushes all agents backward, which is called a chain push. In this case, if the chain of agents approaches a wall, the first blocker is scattered sideways. After a push, SMART replans.

\textbf{Non-king behavior.} If a non-king agent arrives within $5$\,m of the king, it flees away (to the left, right, or in the first non-wall direction). If it is beyond $5$\,m of the king, it follows its own path, treating other agents as dynamic obstacles.

Overall, there are three decoupled layers: (1) RRT* tree- global path via pure pursuit. (2) SMART repair- updates the path when walls/OHZs invalidate edges, and (3) King/flee/push- direct position adaptation.

%%%%%%%%%%%%%%%%%%%%%%%%%%%%%%%%%%%%%%%%%%%%%%%%%%%%%%%%%%%%%%%%%%%%%%%%%%%%%%%%
\section{RESULTS}
This section presents the comparative evaluation results of PECMAN through simulations and experiments. 
\vspace{-6pt}
\subsection{Validation by Simulations}
We evaluate the performance on seven procedurally generated $2$D floorplans. 
Table~\ref{tab:scenarios} provides the scenario specifications.  All scenarios run on an i9-14900K processor with $1{,}000$ trials per scenario per strategy.

\begin{table}[t]
\centering
\caption{Scenario specifications.}\label{tab:scenarios}
\begin{tabular}{lllll}
\toprule
Scenario & Size & Walls & RRT* iters & Agents \\
\midrule
Building     & 32$\times$32\,m  & 41  & 5K   & 2--4 \\
Office       & 32$\times$32\,m  & 42  & 5K   & 2--4 \\
Warehouse    & 32$\times$32\,m  & 28  & 5K   & 2--4 \\
Hospital  & 96$\times$96\,m  & 200 & 30K  & 4 \\
Airport   & 160$\times$160\,m & 250 & 50K  & 4 \\
Campus      & 128$\times$128\,m & 300 & 80K  & 4 \\
University & 192$\times$192\,m & 400 & 160K & 4 \\
\bottomrule
\end{tabular}
\vspace{3pt}
\end{table}

\begin{figure*}[t]
\centering
\includegraphics[width=0.98\textwidth]{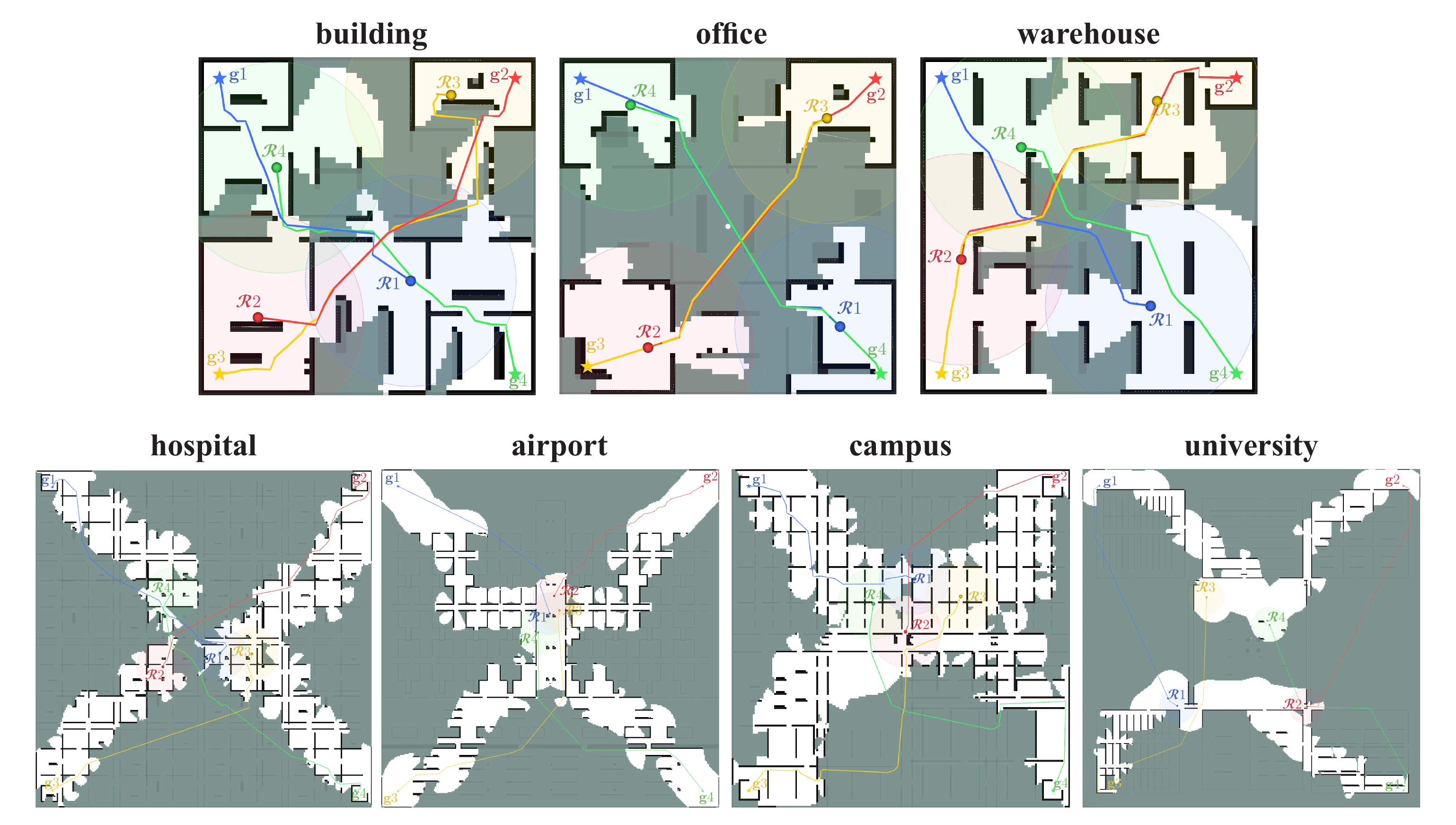}
\caption{$2$D scenarios ranging from $32$\,m$\times$$32$\,m (building, office, warehouse) to $192$\,m$\times$$192$\,m (university). Walls (black) and Unexplored (grey).}
\label{fig:2d_scenarios}
\vspace{-3pt}
\end{figure*}

\begin{figure}[t]
\centering
\begin{minipage}[t]{0.42\textwidth}
\centering
\textbf{(a) Median completion time}\\[0.5mm]
\includegraphics[width=\linewidth]{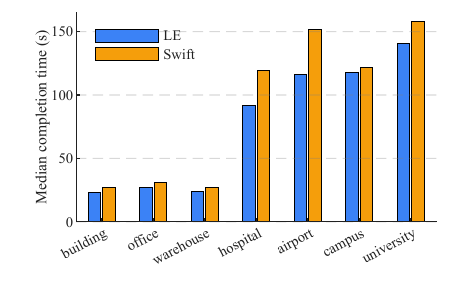}
\end{minipage}\hfill \\
\begin{minipage}[t]{0.42\textwidth}
\centering
\textbf{(b) Average number of rebuilds}\\[0.5mm]
\includegraphics[width=\linewidth]{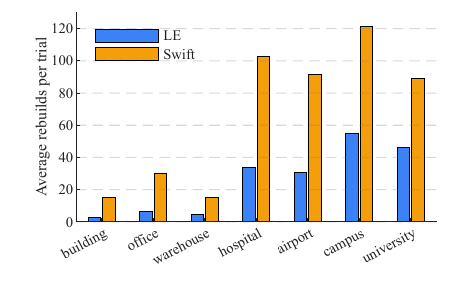}
\end{minipage} \vspace{-12pt}
\caption{Comparison of Lazy Eager vs.\ Swift strategies for $4$-agent simulations with $1{,}000$ trials per scenario.}
\label{fig:e3_senior_benchmarks}
\vspace{6pt}
\end{figure}

\begin{figure}[t]
\centering
\begin{minipage}[t]{0.42\textwidth}
\centering
\textbf{(a) Median completion time}\\[0.5mm]  
\includegraphics[width=\linewidth]{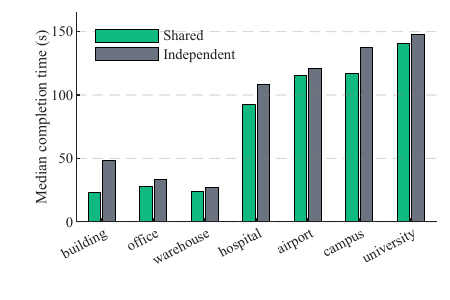}
\end{minipage}\hfill \\
\begin{minipage}[t]{0.42\textwidth}
\centering
\textbf{(b) Fairness gap}\\[0.5mm]
\includegraphics[width=\linewidth]{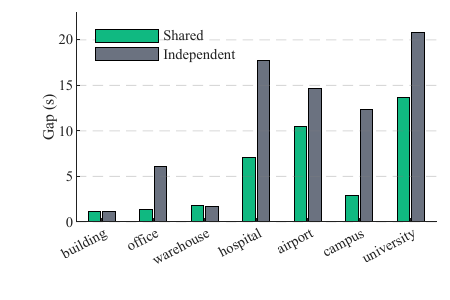}
\end{minipage} \vspace{-12pt}
\caption{Comparison of Shared perception vs Independent modes for $4$-agent simulations with $1{,}000$ trials per scenario.}
\label{fig:sharing_benchmarks}
\vspace{6pt}
\end{figure}

\vspace{6pt}
Fig.~\ref{fig:2d_scenarios} shows snapshots of the multi-agent simulations on seven scenarios: Building, Office, Warehouse, Hospital, Airport, Campus and University. The simulations are conducted in cross mode, where the agents start at the corners and navigate to the opposite corners. The pedestrians follow a random motion at $0.5-2.0$\,m/s.

\vspace{6pt}
%%%%%%%%%%%%%%%%%%%%%%%%%%%%%%%%%%%%%%%%%%%%%%%%%%%%%%%%%%%%%%%%%%%%%%%%%%%%%%%%

\begin{figure*}[!t]
\centering
\includegraphics[width=0.98\textwidth]{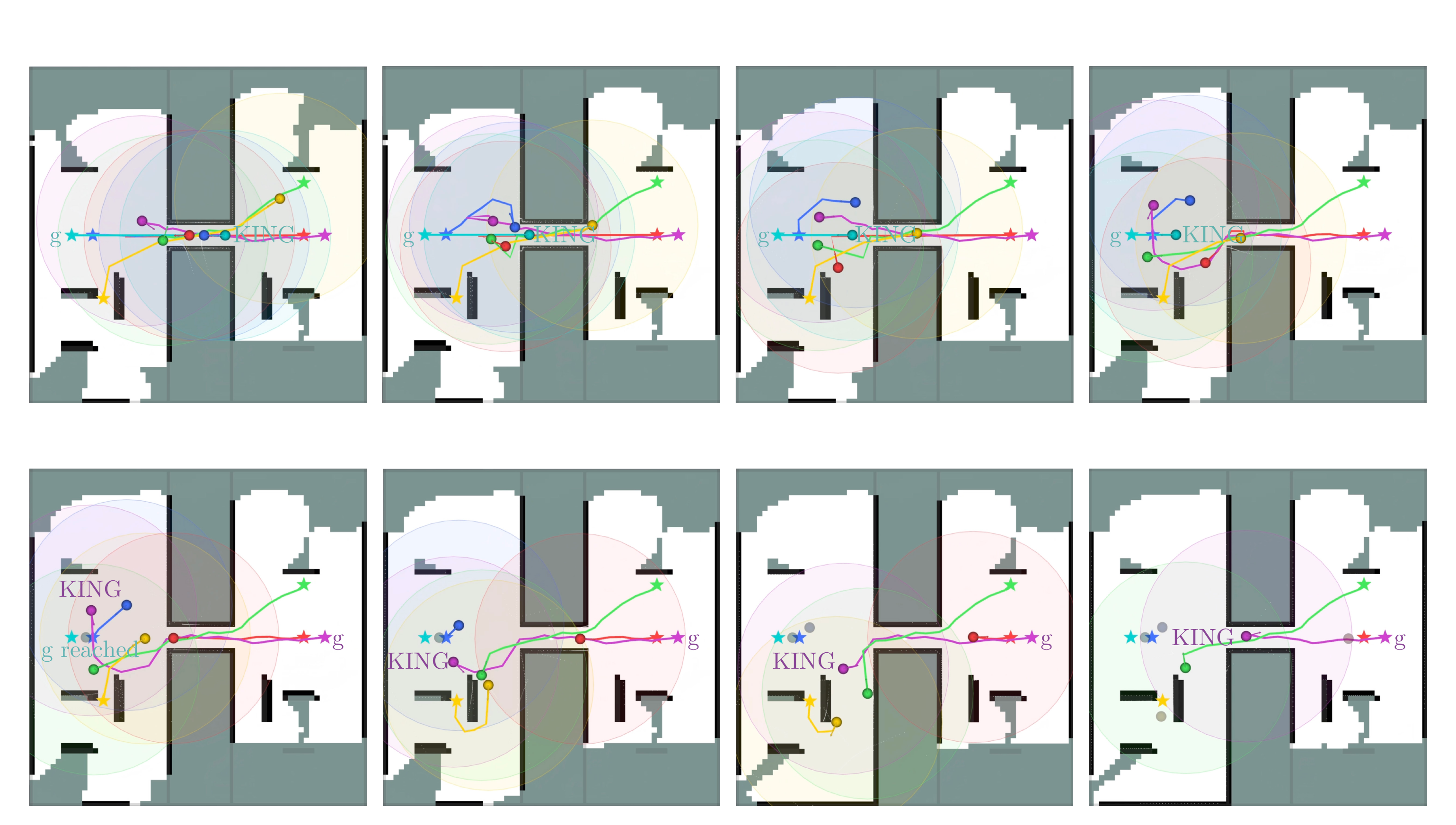}
\caption{Narrow corridor stress test: six agents approach a narrow corridor from both ends under Global King coordination. Eight time-stamped snapshots are shown from one trial. Panels 1--4: the light-blue agent is the active King; the other agents yield by fleeing, being pushed in a chain, or scattering sideways when blocked, allowing the King to pass through the corridor. Panels 5--8: after the previous King reaches its goal, the purple agent becomes the next King; the yellow and green agents yield while it passes.
}
\label{fig:bottleneck_runtime}
\vspace{-6pt}
\end{figure*}

\vspace{6pt}
\subsubsection{Lazy Eager vs. Swift strategies}\label{sec:e3_senior}

We compare two repair strategies: Lazy Eager 
and Swift. As discussed earlier, Lazy Eager scans all edges, but it does that only when the current path is blocked. On the other hand, Swift only scans the edges that are within $10$\,m of the current path. Although both of these methods can leave invalid edges on the map, Lazy Eager has the opportunity to clean them when the path is blocked; thus, it triggers a lower number of full tree rebuilds. Fig.~\ref{fig:e3_senior_benchmarks} visualizes the median completion times and the average number of rebuilds over all trials for each scenario. It is observed that Swift triggers $\sim$$2$--$5\times$ more full rebuilds due to invalid edges. When King pushes an agent into an unexplored region, it encounters invalid edges and must do a full rebuild. This effect is clearer on large maps, e.g., hospital where the rebuild count jumps from $34.0$ to $102.7$ and campus where it jumps from $54.9$ to $121.3$. Lazy Eager delivered a $\sim$$100\%$ success rate, while the team finishes faster in all scenarios. The completion time gap widens on the larger maps, while the rebuild gap is already pronounced even on the small layouts.

\vspace{6pt}
\subsubsection{Shared perception vs. Independent operation modes}\label{sec:sharing}
Next, we compare the performance of the Shared perception and Independent operation modes while using the Lazy Eager repair strategy. In independent operation mode, each agent reacts to an unknown structure when its LiDAR sensor detects it; in contrast, in shared perception mode, a structure discovered by any agent is broadcast to all teammates, who then react accordingly by morphing their respective trees. 

Fig.~\ref{fig:sharing_benchmarks} visualizes the median completion times and the fairness gap over all trials for each scenario. The fairness gap is the difference between the slowest and fastest agent median completion times within each scenario. It is observed that the shared perception mode lowers the team completion time in all scenarios and also decreases the fairness gap in most cases.

The success rate remains $\sim$$100\%$ for both shared perception and independent operation modes in all scenarios. On university, the shared perception mode records an average of $22.8$ rebuilds per trial vs $11.3$ for the independent operation mode, yet it completes slightly faster (140.6\,s vs.\ 148.1\,s). This is because shared perception allows proactive replanning rather than reactive replanning.

\subsubsection{Narrow corridor coordination}\label{sec:deadlock}
The Global King layer resolves the failure mode that appears when two or more agents converge in a corridor narrower than twice the robot radius. In such situations, the reactive methods such as ORCA~\cite{berg2011reciprocal} deadlock because no feasible avoidance velocity exists. As a stress test for this layer, we constructed a $2$\,m wide and $8$\,m long corridor with agents entering from both ends, as shown in Fig.~\ref{fig:bottleneck_runtime}. The King sequencing (i.e., King passes first, others flee; then the next-priority agent becomes the king after the previous one exits) successfully sequenced all robots through the corridor in every trial. It is observed that the King layer handles the corridor case where reactive methods cannot. Fig.~\ref{fig:bottleneck_runtime} shows a representative run where six agents are sequenced through the corridor under King coordination.

\begin{figure}[!t]
\centering
\begin{minipage}[t]{0.48\columnwidth}
\centering
\includegraphics[width=\linewidth]{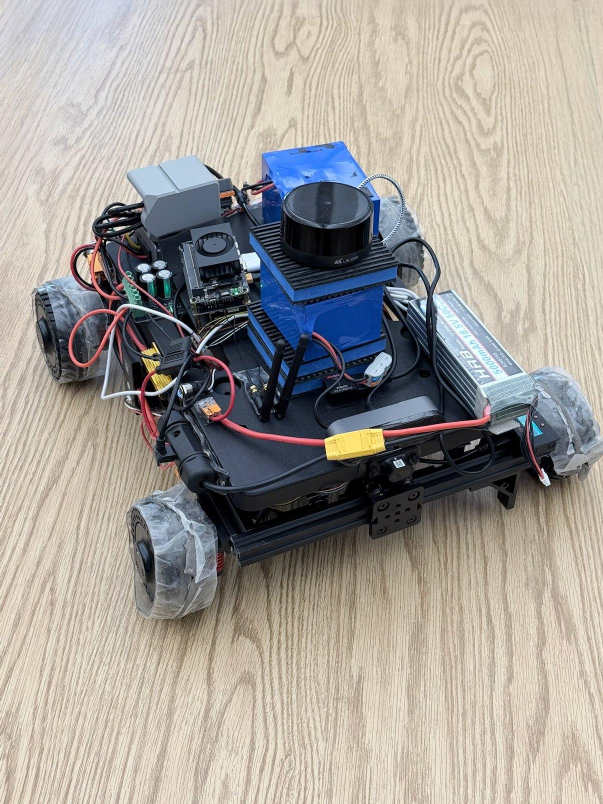}
\\[0.5mm]
\textbf{(a) Robot 1}
\end{minipage}\hfill
\begin{minipage}[t]{0.48\columnwidth}
\centering
\includegraphics[width=\linewidth]{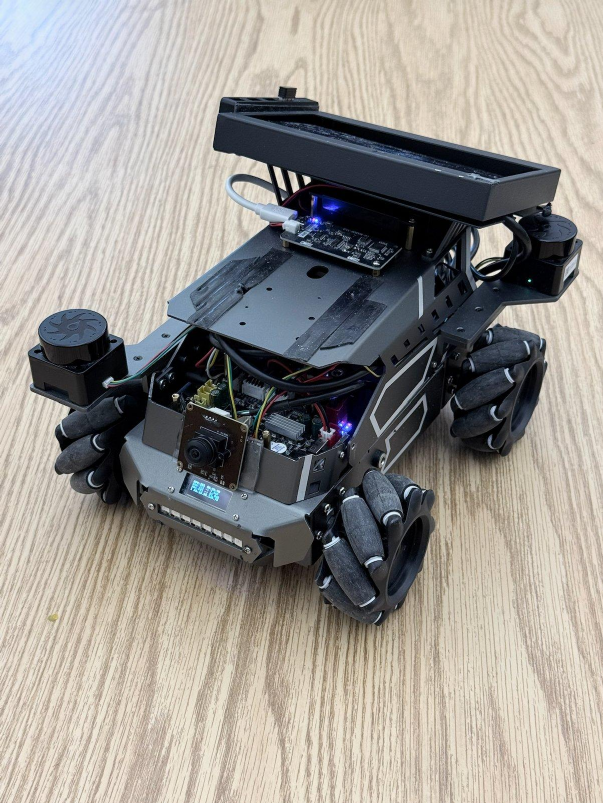}
\\[0.5mm]
\textbf{(b) Robot 2}
\end{minipage}
\caption{Two ROS2-integrated robots used for the experiments. Each robot runs Cartographer SLAM on its onboard Raspberry~Pi~5 and streams observations to the central i9 planner.}
\label{fig:deployment}
\end{figure}

\vspace{-6pt}
%%%%%%%%%%%%%%%%%%%%%%%%%%%%%%%%%%%%%%%%%%%%%%%%%%%%%%%%%%%%%%%%%%%%%%%%%%%%%%%%
\subsection{Testing by Real Experiments}
\label{sec:deployment}

We implemented PECMAN on a team of two heterogeneous robots using ROS2 platform as a final validation. Fig.~\ref{fig:deployment} shows the two physical robots.

\vspace{6pt}
\textbf{Hardware.} Each robot carries a 2D LiDAR and a Raspberry~Pi~5 (4-core, 2.4\,GHz) running Cartographer SLAM, the pose publisher, and a low-level velocity controller. A central Alienware i9-14900K runs  \smarttd{} planner per robot in Rust. ROS2 Humble with CycloneDDS over WiFi connects the two sides. Running Cartographer SLAM on Pi~5 consumes ${\sim}1.5$ cores, while RRT$^{*}$ takes ${\sim}300$\,ms on the Pi~5 versus ${\sim}40$\,ms on the i9. Since running both onboard is not feasible, the i9 hosts the planner while Pi~5 runs the local SLAM, path following controller, and emergency stopping.

\vspace{6pt}
\textbf{Map alignment.} 
Each robot runs SLAM independently, and the i9 aligns the two local maps into a shared planning frame through an external rigid transform that is set from the known deployment geometry at startup. The i9 then transforms occupied cells from each robot's occupancy grid into this shared frame, which becomes the static-obstacle input to \smarttd{}. Thus, map sharing is done at the planner level, and not by merging the underlying SLAM graphs.

\vspace{6pt}
\textbf{Shared perception and execution.} The newly observed static obstacles by one robot are broadcast to the other robot's planner, which updates its tree through incremental repair before it physically encounters that region. The static obstacles are wall portions extracted from the SLAM occupancy map and shared between robots. The dynamic obstacles are distinguished by map consistency. The live LiDAR scan is compared against the currently known static walls. Then, the LiDAR scan points that cannot be explained by the static map are treated as residual points, clustered together (min 8 points, 1.0\,s persistence), and fed to the planner as dynamic obstacles.  The position of the other robot is communicated directly from that robot and not inferred from LiDAR.

\begin{figure}[!t]
\centering
\begin{minipage}[t]{0.48\columnwidth}
\centering
\includegraphics[width=\linewidth]{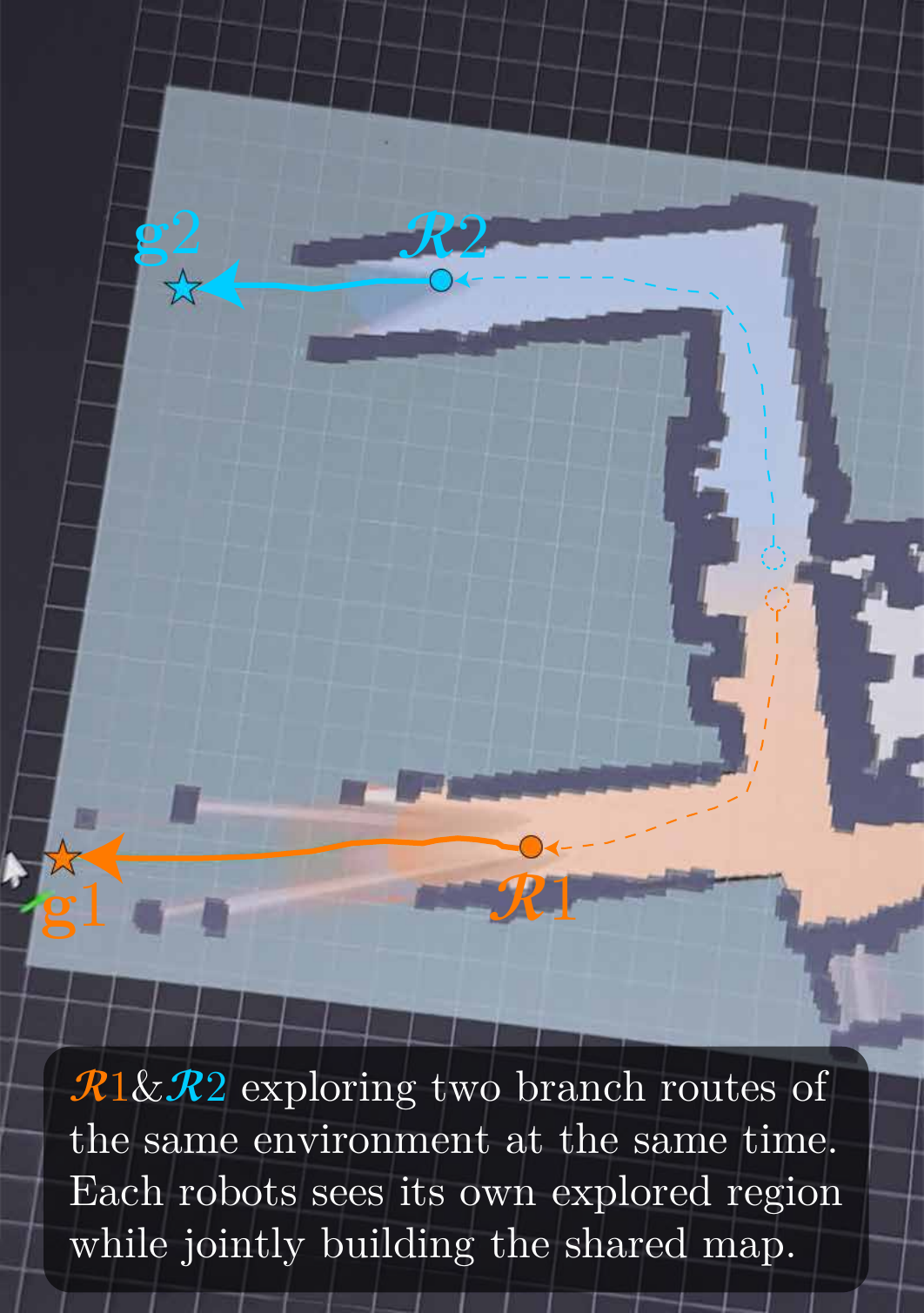 }
\\[0.5mm]
\textbf{(a) Exploration stage}
\end{minipage}\hfill
\begin{minipage}[t]{0.48\columnwidth}
\centering
\includegraphics[width=\linewidth]{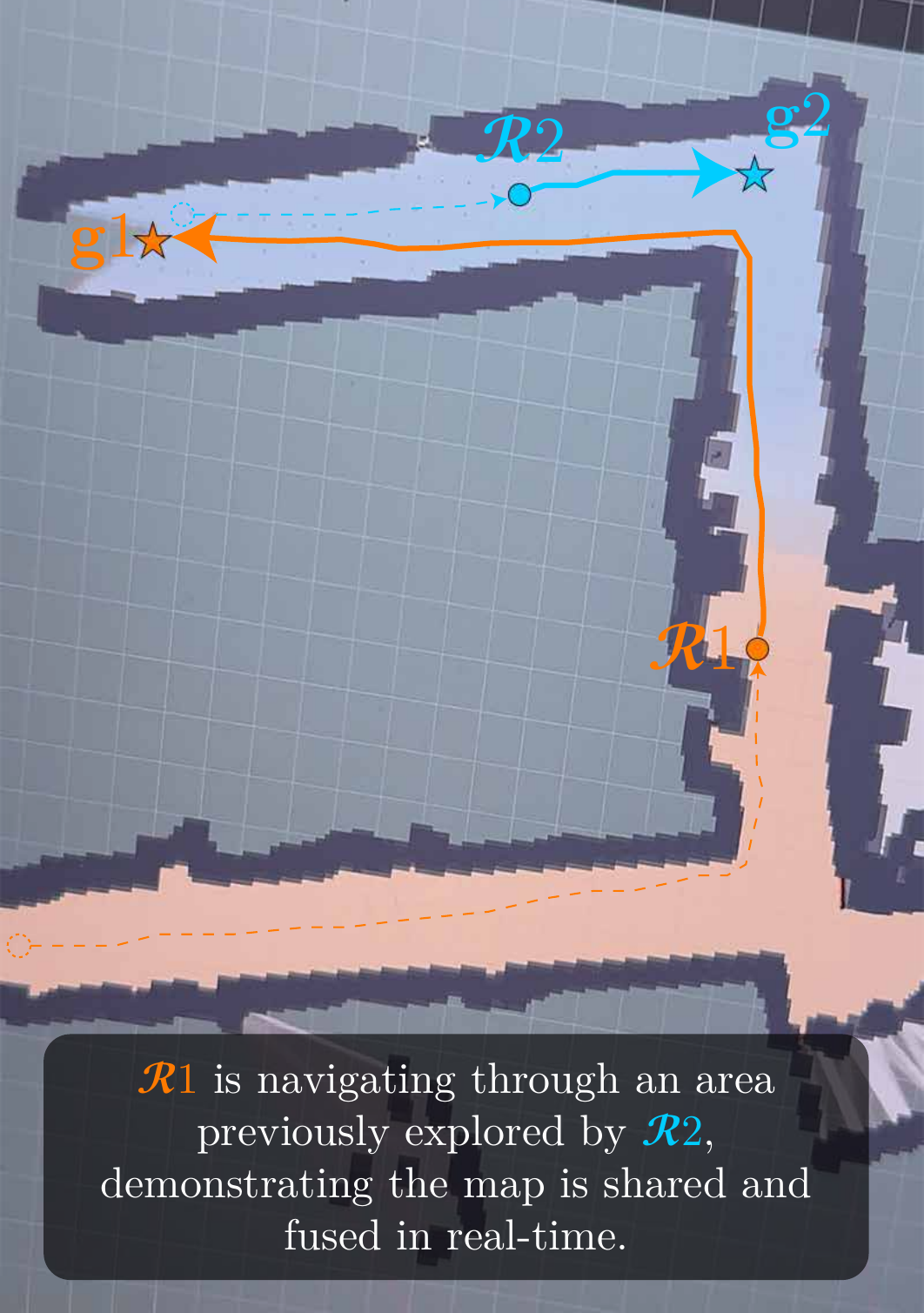}
\\[0.5mm]
\textbf{(b) Tasking stage}
\end{minipage}
\caption{Two-robot team tasking in a U-shaped corridor in a building. (a) Exploration phase- each robot maps its own arm of the corridor; the two partial maps are shared in the planning frame. (b) Tasking phase- Robot 1 (red) finds a path  to a goal in the region explored by Robot 2, while Robot 2 (blue) navigates to a goal in its own region.}
\label{fig:deployment_rviz}
\end{figure}

Fig.~\ref{fig:deployment_rviz} shows a U-shaped corridor exploration in RViz. First, during the exploration stage, each robot explores one arm of the corridor, maps it, and shares it with the other robot. Then, in the  tasking stage, each robot is assigned different goal points such that Robot 1 is assigned a goal in the arm that was scanned by Robot 2, while Robot 2 is assigned a goal in its own arm. It is seen that both Robots 1 and 2 successfully plan their paths to their respective goals and navigate to them, thus confirming the end-to-end success of the shared-perception.

\section{Conclusion}
This paper extends \smarttd{}, which is a single-robot real-time tree-repair planner, to perception-enabled collaborative multi-agent navigation planner (PECMAN). It is shown that PECMAN works in unknown environments through shared perception between agents. This enables agents to conduct proactive replanning in areas not observed by them, thus making more efficient plans and saving computation times. PECMAN was validated via extensive Monte-carlo simulations for seven 2D scenarios. It is seen that Shared perception achieves significant speed ups in the team-completion times, while maintaining near-$100\%$ success rates. Future work involves extending the PECMAN algorithm to consider curvature constrained robots ~\cite{song2019} and environmental currents~\cite{Mittal2020DubinsCurrents}.

\bibliographystyle{IEEEtran}
\bibliography{PECMAN_Arxiv}

\end{document}